\documentclass[conference]{IEEEtran}
\IEEEoverridecommandlockouts    

\usepackage{multirow}
\usepackage{lipsum}
\usepackage{amsmath}
\usepackage{amssymb}
\usepackage{graphicx}
\graphicspath{{./Figures/}}
\usepackage{xcolor}      
\usepackage{tikz}
\usepackage{comment}
\usepackage{graphicx}    
\usepackage{multirow}
\usepackage{booktabs}
\usepackage{pifont}
\usepackage{caption}     
\usepackage{subcaption}  
\usepackage{algorithm}
\usepackage{algpseudocode}
\usepackage{rotating}
\usepackage{url}

\newcommand{\cmark}{\ding{51}}%
\newcommand{\xmark}{\ding{55}}%
\newcommand{\xhdr}[1]{ \noindent {\textbf{#1}}}

\title{\LARGE \bf Towards Driver Behavior Understanding: Weakly-Supervised Risk Perception in
Driving Scenes}

\author{Nakul Agarwal$^{1}$, Yi-Ting Chen$^{2}$, Behzad Dariush$^{1}$}

\begin{document}
	
	\maketitle
    
    \begingroup

\renewcommand{\thefootnote}{\fnsymbol{footnote}}
\footnotetext[0]{$^{1}$Honda Research Institute USA, Inc.}
\footnotetext[0]{$^{2}$National Yang Ming Chiao Tung University.}
\footnotetext{\textsuperscript{$\dagger$}\ \url{https://usa.honda-ri.com/raid}}
\endgroup
	\thispagestyle{empty}
	\pagestyle{empty}
	
	\begin{abstract}
		Achieving zero-collision mobility remains a key objective for intelligent vehicle systems, which requires understanding driver risk perception—a complex cognitive process shaped by voluntary response of the driver to external stimuli and the attentiveness of surrounding road users towards the ego-vehicle. To support progress in this area, we introduce RAID\textsuperscript{$\dagger$} (\textbf{R}isk \textbf{A}ssessment \textbf{I}n \textbf{D}riving scenes) — a large-scale dataset specifically curated for research on driver risk perception and contextual risk assessment. RAID comprises 4,691 annotated video clips, covering diverse traffic scenarios with labels for driver's intended maneuver, road topology, risk situations (e.g., crossing pedestrians), driver responses, and pedestrian attentiveness. Leveraging RAID, we propose a weakly supervised risk object identification framework that models the relationship between driver's intended maneuver and responses to identify potential risk sources. Additionally, we analyze the role of pedestrian attention in estimating risk and demonstrate the value of the proposed dataset. Experimental evaluations demonstrate that our method achieves 20.6\% and 23.1\% performance gains over prior state-of-the-art approaches on the RAID and HDDS datasets, respectively. 
	\end{abstract}
	
	\section{Introduction}
	\label{sec:intro}
    Road traffic accidents remain one of the leading causes of non-natural deaths globally, claiming over 1.3 million lives annually—around 3,700 per day~\cite{who}. Advances in automated and driver assistance systems hold promise for significantly reducing such incidents in future mobility systems. A key research direction for deploying intelligent driving systems is developing computational models of driver decision-making and risk perception in response to surrounding traffic participants. Modeling driver behavior is inherently complex and remains an open challenge, involving both low-level control (steering, braking, throttle) and high-level cognition, such as interpreting other agents’ actions, intentions, and interactions—all shaped by the environment~\cite{Wang_DriverBeahvior2014}. At the cognitive level, drivers first identify relevant elements in the scene, second reason about the interrelationships among these elements, and third infer the future actions of nearby traffic participants~\cite{Endsley_SA_2000,Rasouli_review_ITS2020}.
    
    \begin{figure}[t]
    \centering
    \includegraphics[width=0.45\textwidth,height=6.8cm]{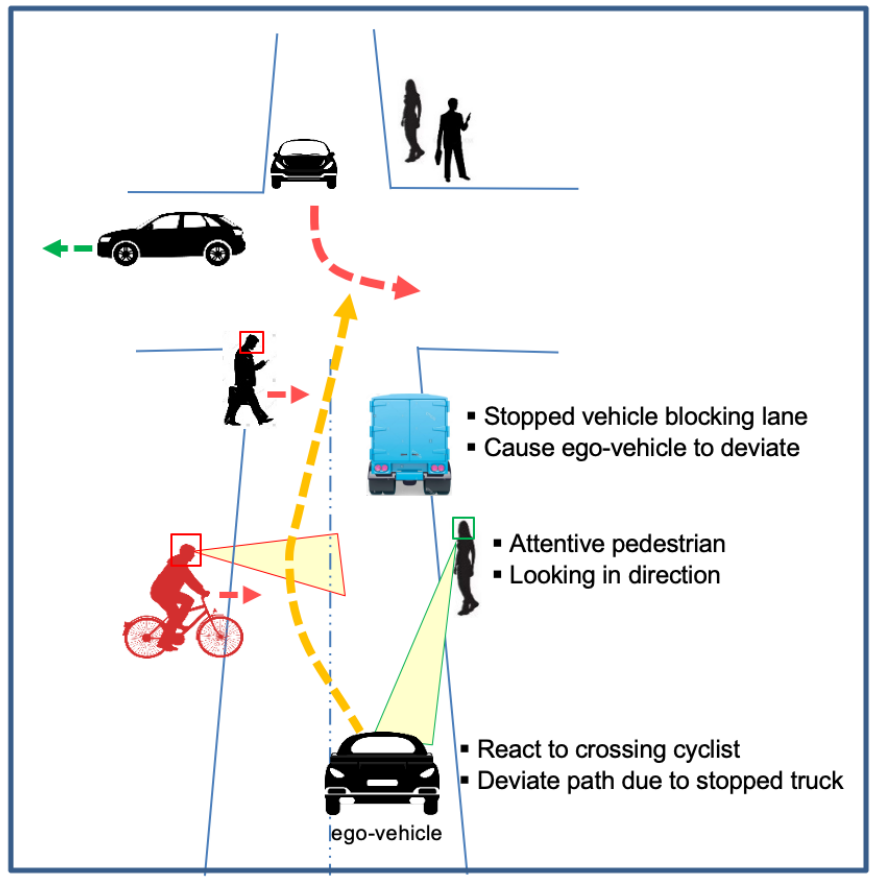}
    \caption{Perception of risk is a complex cognitive process that is manifested, among other things, by a voluntary response of the driver to external stimuli (e.g. deviating from the planned path in response to a truck that is blocking the path) as well as the apparent attentiveness of participants in the scene (e.g. crossing cyclist that is not attentive to the ego-vehicle).}
    \label{fig:introduction}
    \vspace{-1.5em}
    \end{figure}
    
    \begin{table*}[t]
    \small
    \setlength{\tabcolsep}{1.5pt}
    \centering
    \caption{\label{tab:comparison} Comparison of our proposed RAID with other driving scene datasets. '-' indicates the trait is present but not quantified/categorized in the dataset; '\xmark' indicates the trait is absent.}
    \begin{tabular}{c|c|c|c|c|c|c|c|c|c|c|c|c}
                                   & Videos & \begin{tabular}[c]{@{}c@{}}Avg. Video  \\ Duration (s)\end{tabular}  & \begin{tabular}[c]{@{}c@{}} Risk \\ Objects\end{tabular} & \begin{tabular}[c]{@{}c@{}} Risk \\ Situations\end{tabular} & \begin{tabular}[c]{@{}c@{}}Driver \\ Action\end{tabular}  & \begin{tabular}[c]{@{}c@{}}Driver \\ Response\end{tabular}  & \begin{tabular}[c]{@{}c@{}}CAN \\ Data\end{tabular}  & LiDAR & \begin{tabular}[c]{@{}c@{}}Road \\ Topology\end{tabular}  & \begin{tabular}[c]{@{}c@{}}Pedestrian \\ Attention\end{tabular} &  \begin{tabular}[c]{@{}c@{}}Face \\ Annotations\end{tabular}    & \begin{tabular}[c]{@{}c@{}}Pedestrian \\ Tracklets\end{tabular}  \\ \hline
    JAAD~\cite{Rasouli_ICCVW2017}            & 346 & 5-15 & \xmark &\xmark & \xmark & \xmark & \xmark & \xmark & \xmark & \cmark & \xmark & \cmark  \\ 
    PIE~\cite{Rasouli_pie_ICCV2019}            & 53 &  600 &\xmark & \xmark& \xmark & \xmark & \cmark & \xmark & \cmark & \cmark & \xmark & \cmark  \\ 
    STIP~\cite{liu2020spatiotemporal}                & 556 & - &\xmark & \xmark & \xmark & \xmark & \xmark & \xmark & \xmark & \xmark & \xmark & \xmark  \\ 
     LOKI~\cite{girase2021loki}           & 644 & 13 &\xmark & \xmark & \cmark &\xmark  & \cmark & \cmark & \xmark & \xmark & \xmark & \cmark  \\ 
    OATS~\cite{agarwal2023ordered}           & 1026 & 20 &\xmark & \xmark & \cmark &\xmark  & \cmark & \cmark & \xmark & \xmark & \xmark & \cmark  \\ 
    DRAMA~\cite{malla2023drama}           & 17785 & 2 & 17066 & - & \cmark &\xmark  & \cmark & \xmark & \xmark & \xmark & \xmark & \xmark  \\ 
    Rank2Tell~\cite{sachdeva2024rank2tell}           & 116 & 20 & 54116 & - & \cmark &\xmark  & \cmark & \cmark & \xmark & \xmark & \xmark & \cmark  \\ 
    HDDS~\cite{li2023TPAMI}                & - & - & 630 & 4 & \cmark & \cmark & \cmark & \cmark & \xmark & \xmark & \xmark & \xmark  \\ 
     \textbf{RAID}        & 4691 & 7 & 75873 & 10 & \cmark & \cmark & \cmark & \cmark &  \cmark & \cmark & \cmark & \cmark  \\  
    \end{tabular}%
    \end{table*}

    In the first level of the aforementioned cognitive process, drivers aim to avoid risky situations based on their perception of risk~\cite{riskperception}. While intelligent vehicle systems often define risk in terms of collision prediction~\cite{risk_assessment_Lefevre_ROBOMECH}, this does not fully capture how drivers perceive risk. A driver-centric definition of risk was introduced in~\cite{Li_risk_object_IROS2020}, where risk is inferred not by explicit collision probability but from a driver’s behavioral response, such as deviating from the path to avoid a blocking truck (see Figure~\ref{fig:introduction}). Although this formulation initiated the \textit{risk object identification} task in a weakly supervised manner—the focus of our work—prior studies were constrained by limited scenario diversity and lacked key behavioral cues, such as pedestrian attentiveness—an essential factor for comprehensive risk assessment and road safety.
    
    Pedestrian attentiveness is a critical factor in risk perception, as it facilitates non-verbal communication and mutual intention understanding between drivers and pedestrians—key to modeling their interactions~\cite{Rasouli_review_ITS2020}. As illustrated in Figure~\ref{fig:introduction}, joint attention (e.g., a pedestrian waiting to cross making eye contact with the driver) helps reduce uncertainty and fosters mutual awareness. Recent works~\cite{Kooij_pedestrianContext_ECCV2014,Rasouli_ICCVW2017,Rasouli_pie_ICCV2019} have explored pedestrian intention prediction from both algorithmic and dataset perspectives, but with notable limitations. Specifically,~\cite{Rasouli_ICCVW2017,Rasouli_pie_ICCV2019} ignore whether pedestrians influence the driver's decision-making or motion planning. Although pedestrians' attention is annotated, it is not linked to downstream tasks. Furthermore, these datasets lack head annotations—an important cue for assessing attentiveness (see Section~\ref{results}). The dataset in~\cite{Kooij_pedestrianContext_ECCV2014} is collected in controlled settings, where the pedestrian is always the sole relevant agent, limiting real-world applicability. To advance research in this area, more realistic and task-driven datasets are needed. To our knowledge, RAID is the first large-scale dataset in naturalistic driving scenes that includes diverse risk situations, pedestrian attention and face annotations in the context of risk perception (see Table~\ref{tab:comparison}. We also introduce simple baselines to benchmark RAID.
    
    To this end, the contributions of this paper are summarized as follows. 
    \textbf{First}, to address the limitations of existing datasets, we introduce a novel and comprehensive dataset with a diverse set of risk situations and annotations to enable research for risk object identification. \textbf{Second}, we provide a weakly supervised model for risk object identification which models the relationship between driver's action and response, and benchmark it on the proposed dataset along with other baselines for future research. \textbf{Third}, we provide pedestrian attentiveness labels on a subset of the proposed dataset and a face-based attention detection method to enable joint study of pedestrian attention and risk perception—an aspect unexplored in prior work.
	
	\section{Related Work}
	\label{sec:formatting}

    \xhdr{Risk Object Identification.}
    Risk object identification aims to identify objects influencing driver behavior~\cite{Li_risk_object_IROS2020}. Existing approaches fall into four categories, each with key limitations. First, pixel-level driver attention prediction by imitating gaze~\cite{Alletto_Dreye_cvprw2016,xia2019predicting,tawari2018learning}, although gaze is noisy and often unrelated to influential objects. Second, risky region localization~\cite{Zeng_agentrisk_cvpr2017} or object importance estimation (binary classification)~\cite{gao2019goal, li2022important}, which rely on subjective, annotation-heavy supervision and fail in unseen scenarios. Third, attention maps from end-to-end driving models~\cite{kim2017interpretable,wang2019deep}, but these may not align with truly causal objects~\cite{Haan_causalconfusion_nips2019}. Fourth, weak supervision from driver response to identify risk objects~\cite{Li_risk_object_IROS2020, li2023TPAMI, agarwal2023driver}. Our approach most closely aligns with this set of approaches and also uses driver response but differs by i) effectively modeling traffic agents and interactions (Table~\ref{comparison_hdd} and Table~\ref{comparison_roi}), ii) jointly considering pedestrian attentiveness for risk assessment, and iii) validating on multiple datasets, including our diverse RAID dataset with many more risk situations.

    \xhdr{Pedestrian Behavior Understanding.} Pedestrian behavior understanding has been studied with a long history.
    Pedestrian detection is an essential first step towards modeling pedestrian behavior.
    A collective effort in establishing common benchmarks has stimulated the progress of pedestrian detection algorithms~\cite{Zhang_filteredchannel_cvpr2015,Wang_repulsion_cvpr2018,Zhang_OR_eccv_2018,luo_multimodal_cvpr2020}.
    To estimate motion of pedestrians, pedestrian tracking algorithm~\cite{Bera_adapt_icra2014, Rezatofighi_iccv2015,Wojke2018deep,Bergmann_tracking_iccv2019} and the corresponding benchmarks~\cite{Geiger2012CVPR,mot16} are introduced.
    To realize proactive warning systems, pedestrian motion forecasting are  investigated~\cite{Kooij_pedestrianContext_ECCV2014,Alahi_sociallstm_cvpr2016,Gupta_socialgan_cvpr2018,Malla_TITAn_cvpr2020,dax2023disentangled, chi2023adamsformer}.
    In addition, pedestrian intention prediction~\cite{Kooij_pedestrianContext_ECCV2014,Rasouli_ICCVW2017,Rasouli_pie_ICCV2019,liu2020spatiotemporal}, pedestrian action recognition and detection~\cite{Rasouli_pedetrianaction_bmvc2018,Malla_TITAn_cvpr2020}, and pedestrian head orientation estimation~\cite{Hamaoka_headturning_ivw2013,Kooij_pedestrianContext_ECCV2014,Rasouli_ICCVW2017} are studied.
    While different directions have been explored, the interconnection of pedestrian attentiveness and driver behavior in the context of risk assessment is rarely studied, especially using faces.

    \noindent \textbf{Datasets.} 
    As shown in Table~\ref{tab:comparison}, few datasets focus on risk assessment in driving scenes. Rank2Tell~\cite{sachdeva2024rank2tell} and DRAMA~\cite{malla2023drama} include risk objects but lack structuring by ‘Risk Situations’. Since their papers omit risk object counts, we manually compute them as 17,066 (DRAMA) and 54,116 (Rank2Tell). 
    HDDS~\cite{li2023TPAMI} is closest to RAID but diverges in key ways: it has fewer risk situations, about $100\times$ fewer risk object annotations, and no pedestrian attention annotations in the context of risk perception. Moreover, HDDS reports statistics only in annotated frames and not videos (see Table 1 in~\cite{li2023TPAMI}), so we mark video-level statistics as '-' in Table~\ref{tab:comparison} and instead compare on annotated frames—their own reporting metric. 
	
	\section{RAID Dataset}
    \noindent \textbf{Data Collection.} The data is collected in the San Francisco Bay Area using an instrumented vehicle equipped with three Point Grey Grasshopper video cameras with a resolution of 1920 × 1200 pixels, a Velodyne HDL-64E S2 LiDAR sensor, and high-precision GPS and Vehicle Controller Area Network (CAN) systems. CAN data captures driver actions such as steering, braking, and throttle usage. All sensor streams are synchronized and timestamped using ROS along with custom hardware and software, resulting in a unified 10Hz frequency—ensuring each video frame aligns with corresponding GPS, CAN, and LiDAR data.
    
    \noindent \textbf{Annotation Overview.} The video data (10Hz) is first annotated by two external independent annotators—experienced drivers with over five years of driving experience and familiarity with U.S. traffic rules and signage. An internal expert then reviews and finalizes the annotations.
    To assess annotation reliability, we compute the intra-class correlation coefficient (ICC)~\cite{shrout1979intraclass}, a standard metric for inter-rater agreement\cite{Rasouli_pie_ICCV2019}. Our annotations achieve an ICC of 0.87, indicating a high level of consistency (with $\text{ICC} = 1$ denoting perfect agreement~\cite{cicchetti1994guidelines}).
    
\begin{figure}[t]
    \centering
    \includegraphics[width=0.47\textwidth,height=5.2cm]{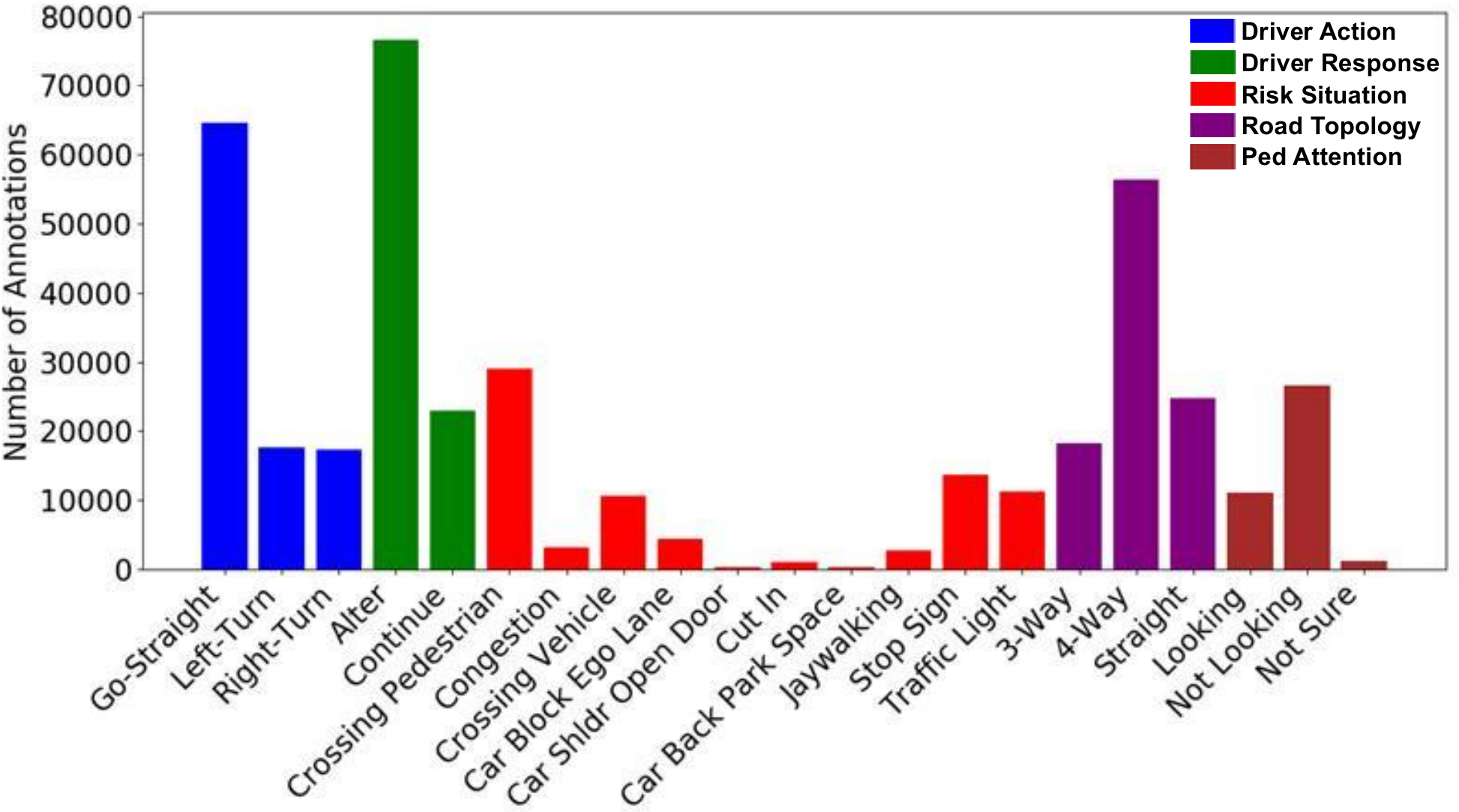}
    \caption{Annotation statistics of our RAID dataset.}
    \label{stats}
    \vspace{-15pt}
    \end{figure}

    \noindent \textbf{Annotation Procedure.} We propose a 4-layer representation—\textbf{Driver Action}, \textbf{Road Topology}, \textbf{Risk Situation}, and \textbf{Driver Response}—to describe driver behavior for risk assessment, with detailed statistics shown in Figure~\ref{stats}. Annotators watch the video data and are instructed to imagine themselves as the driver, identifying situations that cause the ego vehicle to stop or deviate. The \textit{most} influential traffic agent responsible for this behavior is annotated with a bounding box under the \textbf{Risk Situation} layer and categorized into one of ten predefined classes (Figure~\ref{stats}). To aid annotation, ego-vehicle speed from CAN bus data is overlaid on the video to help infer motion states (e.g. stopping). Video clips are formed around these risk situations, each containing both \textit{'Continue'} motion—uninterrupted ego vehicle movement—and \textit{'Alter'} motion—stop or deviate due to a risk object (e.g. deviating around a parked vehicle, stopping for crossing pedestrian, or stop sign), which forms the \textbf{Driver Response} layer. Clips end when the risk object no longer influences the ego vehicle, e.g., after a jaywalker has finished crossing. This results in 4691 clips, each capturing a \textit{single} risk situation. We focus on three classes for \textbf{Driver Action}, i.e., \textit{Left-Turn}, \textit{Right-Turn} and \textit{Go-Straight}. This layer denotes the driver’s eventual intended maneuver, and is annotated as \textit{Go-Straight} until the driver initiates a reaction to the risk object. We also annotate \textbf{Road Topology}, categorized as \textit{4-Way}, \textit{3-Way}, and \textit{Straight}. We note that samples of \textit{Car on Shoulder Open Door} (90) and \textit{Car Back Park Space} (82) are less frequent in RAID due to their rarity in real driving, yet their counts are comparable to HDDS~\cite{li2023TPAMI} (84-311). We retain them in the dataset for zero/few-shot learning, rare-event evaluation, and to capture the long-tail distribution of real-world risk. While RAID exhibits class imbalance reflecting real-world frequency, we consider this an opportunity for future work on modeling rare yet safety-critical events underrepresented in existing datasets.
     

    
    \begin{figure*}[t]
    \centering
    \includegraphics[width=0.87\textwidth,height=5.2cm]{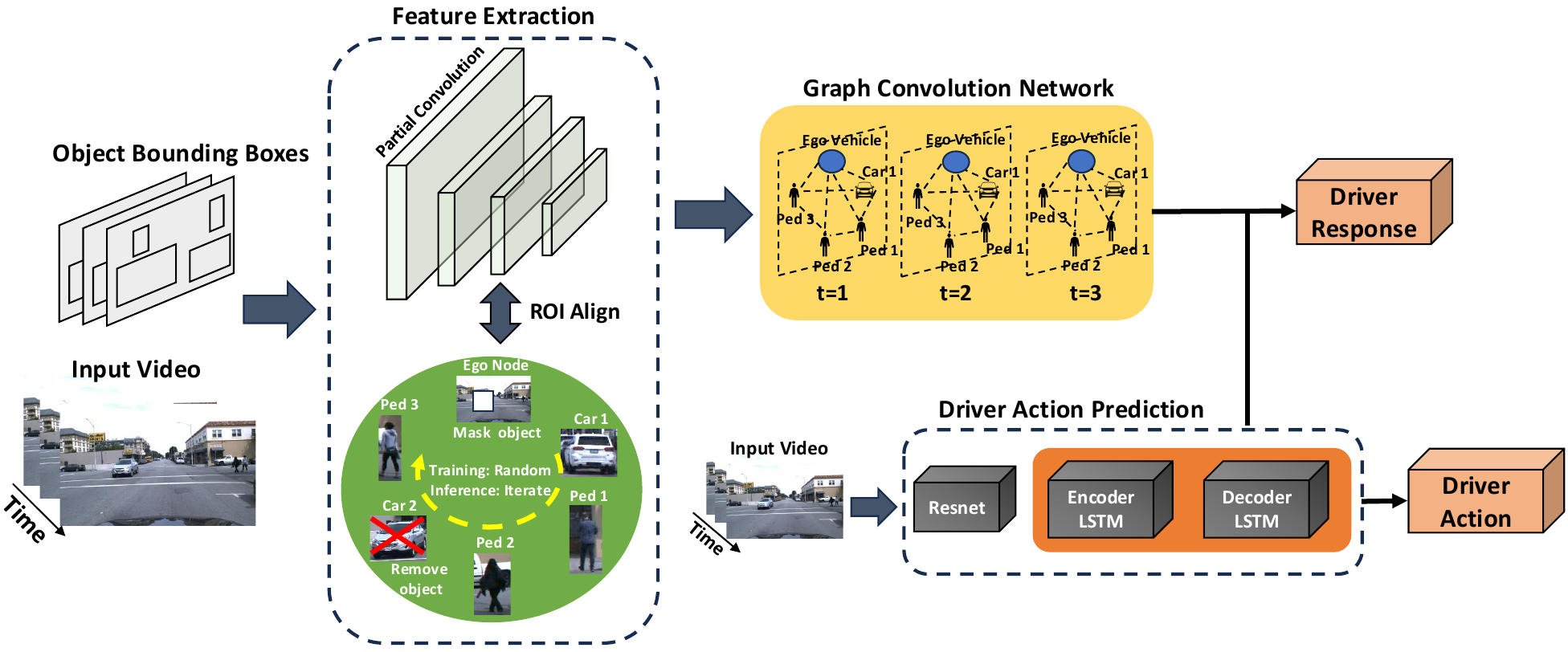}
    \caption{\textbf{Proposed Network Architecture.} The algorithm takes as input a sequence of RGB frames and object tracklets. We then extract corresponding agent-level features using partial convolution and RoIAlign, where in each iteration of the network, the partial convolution removes an agent using a binary mask. These features then form nodes of the graph convolution network. In parallel, the RGB frames are also used to obtain driver's action using a temporal encoder-decoder LSTM network. Finally, the feature representation from the graph and driver's action are combined to predict the driver's response.}
    \label{fig:model}
    \vspace{-1.0em}
    \end{figure*}
    
    For pedestrian attentiveness, we focus on a subset of 695 scenarios from the larger set used for the risk object identification task, where the ego-vehicle is approaching and the driver is influenced—directly or indirectly—by pedestrians. 
    We manually annotate full-body and face bounding boxes along with occlusion flags, allowing attentiveness reasoning based on both facial and body cues, rather than relying solely on body pose~\cite{Rasouli_ICCVW2017,Rasouli_pie_ICCV2019}, as supported in Section~\ref{results}. Tracking information is also provided for these annotated pedestrians.
    Annotation guidelines require pedestrians to be taller than 70 pixels, with face boxes labeled if visible and over 10 pixels. Occlusion is marked as partial (25–75\%) or full ($>$75\%). We use three attentiveness labels: \textit{Looking}, \textit{Not Looking}, and \textit{Not Sure}, with the last capturing cases where gaze interpretation is uncertain due to small or obscured faces.
	
	\section{Methodology}
    \subsection{Risk Object Identification}
    \xhdr{Problem Formulation.} We formulate risk object identification as a cause and effect problem~\cite{Li_risk_object_IROS2020, li2023TPAMI}. Specifically, a two-stage framework is used to identify the cause (i.e., the risk object) of an effect (i.e., driver behavior change). Given an observed video $V_{1,Z} = [F_1, ..., F_Z]$ starting from time $T = 1$ and of length $Z$ where $F_t$ denotes the frame at time $t$, our goal is to identify the object which \textit{most} influences the ego-vehicle in $F_Z$. Our data is clustered into two categories: ‘Continue’- the ego vehicle moves uninterrupted and ‘Alter’- the ego vehicle stops or deviates for traffic agents. We use the behavioral change between these two output states of our model to identify the agent that \textit{most} influences driver behavior. Note that this is the only supervision signal to the model.
    The overall pipeline is shown in Figure~\ref{fig:model} and described in detail below. We benchmark RAID by providing a weakly supervised model inspired by~\cite{li2023TPAMI}, but with some changes in the graph formulation and driver action prediction as described below. We adopt a graph-based model for fair comparison and as a baseline, leaving advance architectures (e.g., transformers) for future exploration.
    
    \xhdr{Graph Definition.} Given a sequence of RGB frames $V_{1,Z} = [F_1, ..., F_Z]$, we detect and track traffic agents using Mask R-CNN~\cite{he2017mask} (pretrained on COCO dataset~\cite{lin2014microsoft}) and Deep SORT~\cite{wojke2017simple}, respectively. We then extract agent-level features using RoIAlign~\cite{he2017mask} to construct a spatio-temporal graph $G_t = (M_t, A_t)$, where $M_t = \{m_t^i |\forall i\in \{1,....,N\}\}$ is the set of vertices and $A_t = \{a_t^{ij}|\forall i,j\in \{1,....,N\}\}$ is the adjacency matrix $\forall t\in \{1,....,Z\}$. The $N$ nodes represent traffic agents in the scene, including the ego vehicle, whose feature is extracted using a frame-sized bounding box to capture global scene context. We restrict traffic agent classes to: \textit{person}, \textit{bicycle}, \textit{car}, \textit{motorcycle}, \textit{bus}, \textit{truck}, \textit{traffic-light}, and \textit{stop-sign}.
    
    Building on the success of appearance features in~\cite{wu2019learning, li2023TPAMI}, we integrate them into our graph formulation. However, unlike them,
    we explicitly account for the variability of object presence caused by tracking inconsistencies or dynamic agent entry/exit within a clip. This enhances both robustness and temporal alignment in graph construction. In our graph, $a_t^{ij}$ models the pairwise relation between two agents at time $t$ and is formally defined as:
    \begin{equation}
        a_t^{ij} = \frac{f_p(m_t^i,m_t^j)\text{exp}(f_a(m_t^i,m_t^j))}{\sum_{j=1}^{N} f_p(m_t^i,m_t^j)\text{exp}(f_a(m_t^i,m_t^j))}, 
    \end{equation}
    
    where $f_a(m_t^i,m_t^j)$ focuses on modeling the appearance relation between agents $i$ and $j$ at time $t$, and $f_p(m_t^i,m_t^j)$ is an indicator function which handles object presence variability. The purpose of $f_a$ is to provide rich semantic context for accurately characterizing the nature of interactions between agents. We use softmax function to normalize the influence on agent $i$ from other agents. The appearance relation is calculated as below:
    \begin{equation}
        f_a(m_t^i,m_t^j) = \frac{\theta(m_t^i)^T\phi(m_t^j)}{\sqrt{D}}
    \end{equation}
    
    where $\theta(m_t^i) = \textbf{w}m_t^i$ and $\phi(m_t^j)=\textbf{w'}m_t^j$. Both \textbf{w}$\in \mathbb{R}^{D\times D}$ and \textbf{w'}$\in \mathbb{R}^{D\times D}$ are learnable parameters which map
    appearance features to a subspace and enable learning the correlation between two objects, and $\sqrt{D}$ is a normalization factor. 
    We define $f_p$ as follows:
    \begin{equation}
    \label{presence}
    f_p(m_t^i,m_t^j) = \mathbb{I}(m_t^i=\text{present} \textbf{ and } m_t^j = \text{present})
    \end{equation}
    
    where $\mathbb{I}(\mathord{\cdot})$ is the indicator function. Once the graph is built, we reason over it using Graph Convolutional Network (GCN)~\cite{kipf2016semi}. GCN takes a graph as input, performs computations over the structure, and returns  
    output relational feature $g$ which is passed through a MLP classifier to obtain driver response score $s$, followed by a standard cross entropy loss. 
    We use partial convolution layer~\cite{liu2018image} to simulate the absence of traffic agents and predict the corresponding driver response~\cite{Li_risk_object_IROS2020, li2023TPAMI}. During inference, agents are iteratively masked, and the one yielding the highest \textit{Continue} confidence is identified as the risk object. 

    \begin{algorithm}[t]
    \caption{Workflow of Driver Action Prediction}\label{alg:cap}
    \hspace*{\algorithmicindent} \textbf{Input:} Sequence of RGB frames $V_{1,Z}$ \\
    \hspace*{\algorithmicindent} \textbf{Output:} Driver action score $p_{act}$ and hidden state $h_e$ 
    \begin{algorithmic}[1]
    \State $x$ := ResNet($V_{1,Z}$).
    \State Initialize $h_e$, $c_e$ and $\Hat{x}$ with zeros.
    \For {$t=1:p_{e}$}
    \State  $h_e^{t+1}, c_e^{t+1}, s_e^{t+1} := E(x^t, \Hat{x}^t, h_e^t, c_e^t)$
    \State  Initialize $h_d$, $c_d$, $f_d$ with zeros.
    \For {$k=1:p_{d}$}
    \State $h_d, c_d, s_d := D(f_d, h_d, c_d)$
    \State $f_d := MLP(s_d)$
    \State $\Hat{x} := \Hat{x} + MLP(h_d)$
    \EndFor
    \State $\Hat{x} := \frac{\Hat{x}}{p_d}$
    \EndFor
    \State $p_{act}$ := Softmax(MLP($h^t_e$))
    \end{algorithmic}
    \label{algo_intention}
    \end{algorithm}

    \xhdr{Predicting Driver Action.} 
    Unlike~\cite{li2023TPAMI} which simply uses an I3D feature extractor and a cross-entropy loss, we model driver action using a more structured temporal encoder-decoder framework, enabling richer temporal reasoning.
    We pass $V_{1,Z}$ through ResNet-50 pre-trained on driving scene dataset~\cite{porzi2019seamless} to obtain image features $x$. These features are then passed through a temporal encoder-decoder network, where the main idea is to train a decoder $D$ that predicts the driver action several frames into the future, and then uses that prediction in the encoder $E$ to classify the driver's intention in the current frame. Both $E$ and $D$ use LSTMs~\cite{hochreiter1997long} as the backbone followed by an MLP layer for classification, although other temporal models such as GRUs~\cite{chung2014empirical} or TCNs~\cite{lea2017temporal} can also be used. The algorithm is described in Algorithm~\ref{algo_intention}. 
    
    More specifically, the decoder $D$ works sequentially to predict driver action for the future frames and their corresponding hidden states \{$h^{0}_d, h^{1}_d,...,h^{p_d}_d$\}, cell states \{$c^{0}_d, c^{1}_d,...,c^{p_d}_d$\} and confidence score \{$s^{0}_d, s^{1}_d,...,s^{p_d}_d$\} for next $p_d$ time steps, where $h^i_t$ for $i\in [0,p_d]$ indicates the hidden state at the $i$-th time steps after $t$. It also learns a future feature representation $\Hat{x}$ by embedding the hidden state vector $h_d$ gathered from all decoder steps $p_d$,
    \begin{equation}
      \Hat{x} = \text{ReLU}(\textbf{W}^T_dh_d + \textbf{b}_d)  
    \end{equation} 
    where $\textbf{W}_d$ and $\textbf{b}_d$ are the parameters of MLP in $D$ used for classification. On the other hand, encoder $E$ takes the previous hidden state $h^{t-1}_e$ as well as the concatenation of the image feature $x^t$ and the predicted future feature $\Hat{x}^t$, and updates its hidden states $h^t_e$ for $p_e$ time steps. It then calculates a softmax distribution $p_{act}$ over candidate classes for driver action. The overall loss function for the encoder-decoder network is given as,
    \begin{equation}
        \gamma = \sum_t (\text{CE}(p^t_{int}, l^t) + \sum_{i=0}^{p_d} \text{CE}(\Hat{p}^i_t, l_{t+i}))
    \end{equation}
    where $\Hat{p}^i_t$ indicates the driver action probabilities predicted by the decoder for step $i$ after time $t$, $l^t$ represents the ground truth, CE denotes cross-entropy loss. Finally, we concatenate hidden state of encoder $h_e$ with graph relational feature $g$ to obtain driver response.

    \subsection{Pedestrian Attentiveness}
    Prior works in this area formulate this as a classification problem and provide bounding boxes of the person as a ground truth~\cite{Rasouli_ICCVW2017,Rasouli_pie_ICCV2019}. While the full body pose gives a cue for attention classification,~\cite{Rasouli_ICCVW2017} reaffirms our intuition by showing that a cropped image around the head acts as a stronger cue for this task. However, since the JAAD dataset introduced in~\cite{Rasouli_ICCVW2017} lacks head annotation, the authors use a heuristic to crop the area around the head. This only offers slight improvement in the results. In contrast, our dataset provides a bounding box around both the faces and body and we use these annotations to tackle this problem from both classification and detection perspective.
    
    For classification, we train cropped images of pedestrians and their faces separately (with minor occlusions up to 25\% allowed) on a ResNet-101~\cite{he2016deep} model and show the advantage of face annotations through the proposed dataset. For detection, we modify the face detector~\cite{deng2019retinaface} by adding a separate head and loss function for estimating pedestrian attention in parallel with box classification and regression branches. More specifically, for any training anchor $i$, we minimize the following multi-task loss:
    \begin{equation}
        \mathcal{L}_{p} = \mathcal{L}_{cls}(p_i,p_i^*) + \mathcal{L}_{box}(t_i,t_i^*) + \alpha\mathcal{L}_{attn}(a_i,a_i^*),
    \end{equation} 
    where $\mathcal{L}_{cls}$ and $\mathcal{L}_{box}$ are face classification and box regression losses similar to~\cite{deng2019retinaface}, $\mathcal{L}_{attn}$ is the loss for the attention head and $\alpha$ is used to balance the attention loss. We use a cross entropy loss for $\mathcal{L}_{attn}$ where $a_i$ is the predicted probability of anchor $i$ corresponding to \textit{looking}, and is non-zero if anchor $i$ is a positive anchor, i.e. has an overlap with the ground truth face box above a threshold $\lambda$. Correspondingly, $a_i^*$ is 1 when the label is \textit{Looking} and 0 if it's \textit{Not Looking}. 
    
    \subsection{Joint Risk Assessment}
    \label{risk_holistic}
    Analysing the role of pedestrian attention in the context of risk for autonomous vehicles (AVs) is challenging for several reasons. In addition to lack of appropriate datasets which focus on scenarios where pedestrian attention is important for driver's risk assessment, establishing a ground truth for attention--such as what the pedestrian is actually focused on--is difficult. Without clear markers, it's hard to assess attention quantitatively or correlate it directly with risk in real-world driving situations. In this work, we primarily focus on the former, and attempt to address the latter by qualitative analysis due to lack of suitable ground truth. We argue that pedestrians looking towards the ego vehicle should have a positive impact on the risk score than the ones who are not, since they are more likely to be attentive towards the car. To this end, we formulate the joint risk as follows:
    \begin{equation}
    \label{joint_risk}
        s_{risk} = \frac{s_{roi}+(1-s_{look})}{2} 
    \end{equation}
    where $s_{look}$ are the logits for \textit{Looking} from pedestrian attention detection framework and $s_{roi}$ corresponds to logits from risk object identification framework. We discuss qualitative results in Section~\ref{results}. Note that we are aware \textit{Looking} is an instantaneous attribute which is hard to link to cognitive awareness of the pedestrian towards ego vehicle. We are also aware that attention may not equally contribute to the joint risk. However, we hope our work will act as a first step in this direction and enable the research community to explore this area.
    \begin{table}[t!]
    \setlength{\tabcolsep}{1pt}
    \caption{\label{comparison_hdd}  Risk object identification results on the HDDS dataset. The best performances are shown in bold. Method with $^\ddagger$ includes driver action.}
    \footnotesize
    \centering
    \begin{tabular}{cccccc}
    \toprule[1pt]
    Method               & \begin{tabular}[c]{@{}c@{}}Crossing \\ Vehicle\end{tabular} & \begin{tabular}[c]{@{}c@{}}Crossing \\ Pedestrian\end{tabular} & \begin{tabular}[c]{@{}c@{}}Parked \\ Vehicle\end{tabular} & \begin{tabular}[c]{@{}c@{}}Cong- \\ estion\end{tabular} & \begin{tabular}[c]{@{}c@{}}Avg \\ mAcc\end{tabular}                 \\ \hline
    Random Selection  &    14.78  &    6.32      &      7.21          &      6.74     &   8.76    \\
     Driver’s Attention~\cite{xia2019predicting}  &    16.80  &    8.90      &      10.00          &      21.30     &   14.25   \\
     Driving Model\cite{Li_risk_object_IROS2020}     &   25.40      &  \textbf{19.88}          &    21.02        &   18.58     &  21.22    \\
     Object-level Attention\cite{wang2019deep}     &  26.52           &    17.50 &  22.20             &     45.05       &   27.81  \\
     DROID~\cite{li2023TPAMI}    & 27.50 & 13.60 & 26.00 & 51.30 & 29.60 \\
     DROID~\cite{li2023TPAMI}$^\ddagger$    & 32.50 & 12.90 & 28.40 & 57.50 & 32.83 \\
    
    Ours                &  \textbf{48.97}  &   18.21   &   \textbf{35.58}   &  \textbf{58.88}  &  \textbf{40.41}     \\   
    \bottomrule[1pt]
    \end{tabular}
    \vspace{-2em}
    \end{table}

    \section{Experiments}
    \label{exp} 
    
    

     \begin{table*}[t!]
    \caption{\label{comparison_roi}  Risk object identification results on the RAID dataset. The best performances are shown in bold.}
    \small
    \centering
    \setlength{\tabcolsep}{3pt}
    \begin{tabular}{cccccccccc}
    \toprule[1pt]
    Method & \begin{tabular}[c]{@{}c@{}}Crossing \\ Pedestrian\end{tabular} & \begin{tabular}[c]{@{}c@{}}Crossing \\ Vehicle\end{tabular} & \begin{tabular}[c]{@{}c@{}}Car Blocking \\ Ego Lane\end{tabular} & \begin{tabular}[c]{@{}c@{}}Cong- \\ estion\end{tabular} & \begin{tabular}[c]{@{}c@{}}Cut-\\ In\end{tabular} & \begin{tabular}[c]{@{}c@{}}Jay-\\ walking\end{tabular} & \begin{tabular}[c]{@{}c@{}}Traffic \\ Light\end{tabular} & \begin{tabular}[c]{@{}c@{}}Stop \\ Sign\end{tabular} & \begin{tabular}[c]{@{}c@{}}Avg \\ mAcc\end{tabular} \\ \hline
    Random Selection& 5.90 & 9.56 & 7.38 & 7.42 & 10.26 & 4.40 & 2.94 & 5.67 & 6.99  \\
    Driving Model~\cite{Li_risk_object_IROS2020} & 13.96 & 24.01 & 15.68 & 30.46 & 33.66 & 11.28 & \textbf{4.32} & 8.00 & 17.67    \\
    Object-level Attention~\cite{wang2019deep} &  13.27 & 24.17 & 14.42 & 41.58 & 32.30 & 12.37 & 2.02 & 6.53 & 18.33    \\ \hline
    Ours  & \textbf{15.64} & 31.05 & \textbf{34.48}  & 27.35  &  27.69 & 3.42   &  2.83  & \textbf{11.76} &  19.28    \\
    Ours+  & 13.89 & \textbf{36.13} & 11.67  & \textbf{54.58}  &  \textbf{38.58} & \textbf{16.06}   &  0.92  & 4.93 &  \textbf{22.10}    \\
    \bottomrule[1pt]
    \end{tabular}
    \vspace{-0.5em}
    \end{table*}
    \subsection{Implementation Details} 
    For HDDS, we use the same train-test split as~\cite{li2023TPAMI}. For RAID, we use eight classes, excluding \textit{Car Backing Into Parking Space} and \textit{Car on Shoulder Open Door} due to limited samples. We apply an 80:20 split on the remaining classes, yielding 3,732 training and 942 testing videos for risk object identification. For pedestrian attention, the data is split into 486 training, 69 validation, and 140 test videos. 
    Our framework is implemented in PyTorch and trained on a single Nvidia V100 GPU. For fair comparison with prior works, the input for HDDS experiments consists of $Z=20$ at 224$\times$224 resolution and 3 fps. For RAID, we set $Z=p_e=p_d=3$ and $N = 25$. The model is trained for 20k iterations with a batch size of 16 using the Adam optimizer (learning rate and weight decay = 0.0005). All model layers are jointly updated during training, except for the pre-trained ResNet-50 used in the driver action module.
    
    For pedestrian's attention classification, we train a ResNet-101 on cropped pedestrian and their face images using SGD (batch size 64, 50 epochs, learning rate 0.001, momentum 0.9). For pedestrian's attention detection, we fine-tune a ResNet-50 pretrained on WIDER FACE~\cite{yang2016wider} using our annotations with SGD (20 epochs, batch size 16, learning rate 0.001, momentum 0.9). The \textit{Not Sure} label is excluded from both pedestrian attention classification and detection tasks.

    \begin{table}[t]
    \caption{\label{tab:other} The average precision (AP\%) results on the different tasks of driver action and driver response.}
    \centering
    \small
    \setlength{\tabcolsep}{1pt}
    \begin{tabular}{ccccccccc}
    \toprule[1pt]
    \multirow{2}{*}{Method} & \multicolumn{4}{c}{Driver Action}  & \multicolumn{3}{c}{Driver Response}  & \multicolumn{1}{c}{Risk} \\
                                            & Left   & Right   & Straight   & mAP &  Continue           & Alter          & mAP & mAcc        \\ \hline
    Ours                                      &      -          &      -       &    -    &   -     &    68.75     &     93.22       &   80.98  &   19.28   \\
    Ours+                                   &    35.32     &     61.48       &   78.96     & 58.59   &        77.34        &     96.43        &   86.88    & 22.10  \\
    \bottomrule[1pt]
    \end{tabular}
    \end{table}
    
    \begin{table}[t!]
    \caption{\label{ped_attn} The average precision (AP\%) of the classification (Cls) and detection (Det) results for pedestrians’ attention.}
    \centering
    \small
    \begin{tabular}{ccccc}
    \toprule[1pt]
                                    & Method & Looking & Not Looking & mAP \\
    \midrule[.5pt]
    \multirow{2}{*}{\rotatebox{90}{Cls}} &   Body~\cite{Rasouli_ICCVW2017}     &   46.08      &      78.14       &  62.10   \\
                                    &   Face     &  \textbf{75.44} & \textbf{92.09} & \textbf{83.76}    \\
    \rotatebox{90}{Det}                       &   Face & 69.83 & 36.03 & 52.93   \\
    \bottomrule[1pt]
    \end{tabular}
    \vspace{-0.5em}
    \end{table}
    
    \subsection{Results and Analysis}
    \label{results}
    \xhdr{Metrics.} 
    We follow the metric from~\cite{Li_risk_object_IROS2020, li2023TPAMI} for risk object identification, where a prediction is correct if its Intersection over Union (IoU) exceeds a certain threshold. We report mean accuracy (mAcc) as the average accuracy across 10 IoU thresholds from 0.5 to 0.95~\cite{zhang2019self}, along with per-class Average Precision (AP) and mean AP (mAP) for overall performance for other tasks.
    
    \xhdr{Baselines.} 
    %
    As very few works address weakly supervised risk object identification, we compare against prior methods relevant to attention modeling and driver behavior–based risk reasoning~\cite{xia2019predicting, Li_risk_object_IROS2020, wang2019deep, li2023TPAMI}.
    While Random Selection and Driver's Attention~\cite{xia2019predicting} are not direct competitors, they serve as reference points for task difficulty. Random Selection chooses a detected object uniformly at random, while Driver’s Attention uses a pretrained gaze model~\cite{xia2019predicting} to select the object with the highest average attention weight, since HDDS lacks gaze labels.
    Among task-relevant baselines,~\cite{Li_risk_object_IROS2020, li2023TPAMI} directly address risk object identification, and~\cite{wang2019deep} is reimplemented and used as an risk object selector based on learned object-level attention.
    As none of these methods release code, reimplementation is required and not all baselines are evaluated on both datasets.
    For HDDS, we report results from~\cite{li2023TPAMI} and reimplement~\cite{wang2019deep, Li_risk_object_IROS2020} to enable consistent evaluation on both HDDS and RAID. Comparison with DROID~\cite{li2023TPAMI} on RAID is omitted due to unavailable code and its complex multi-network design, making reimplementation impractical.

    For pedestrian attention, we use~\cite{Rasouli_ICCVW2017} as a baseline as it explicitly evaluates pedestrian attention in driving contexts. The model is retrained on RAID for fair comparison (Table~\ref{ped_attn}).
    
    %

    \begin{figure*}[ht]
      \centering
    
      \begin{subfigure}[b]{0.45\textwidth}
        \centering
        \includegraphics[width=\linewidth,height=3.6cm]{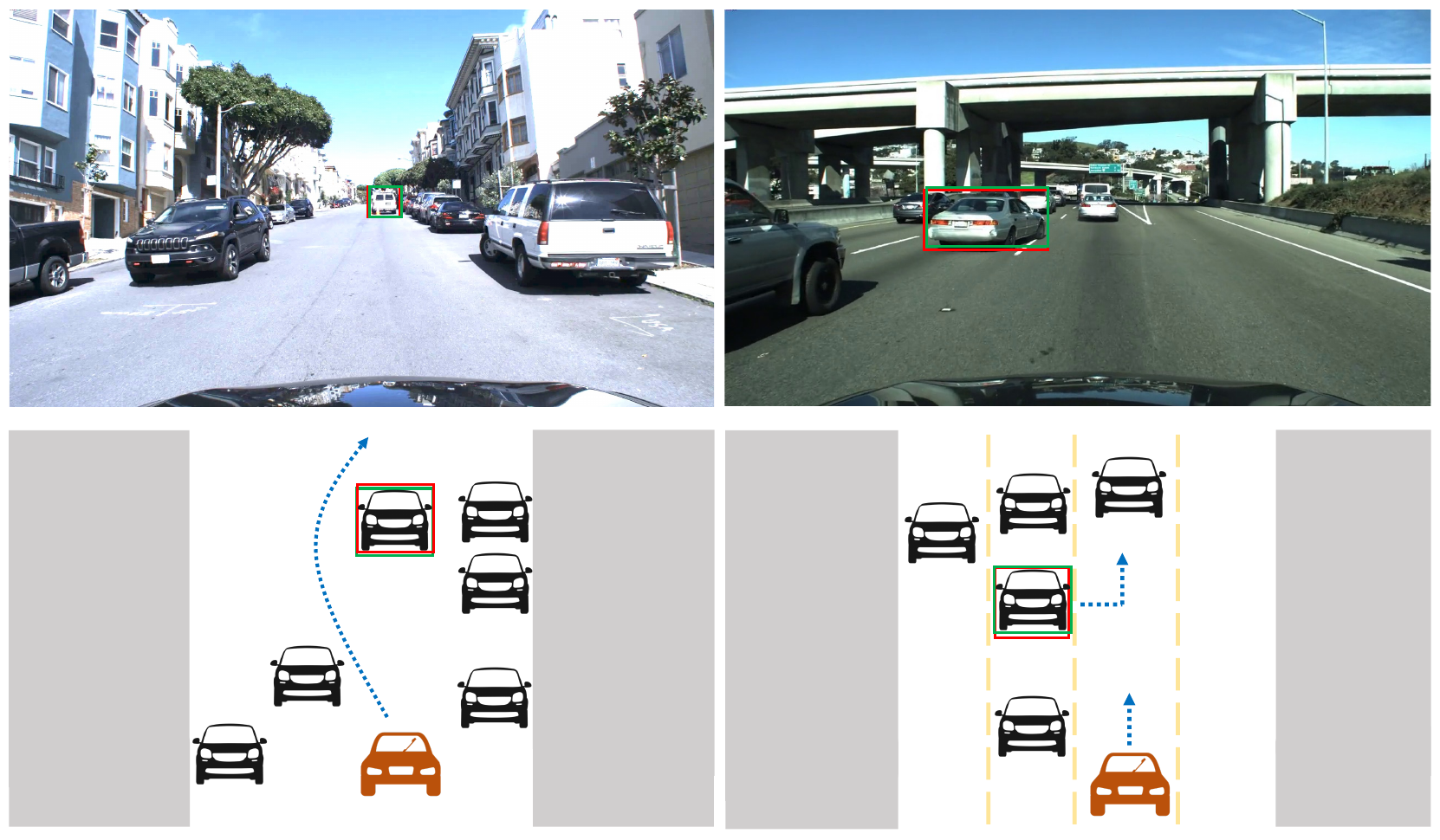}
      \end{subfigure}
      \begin{subfigure}[b]{0.45\textwidth}
        \centering
        \includegraphics[width=\linewidth,height=3.6cm]{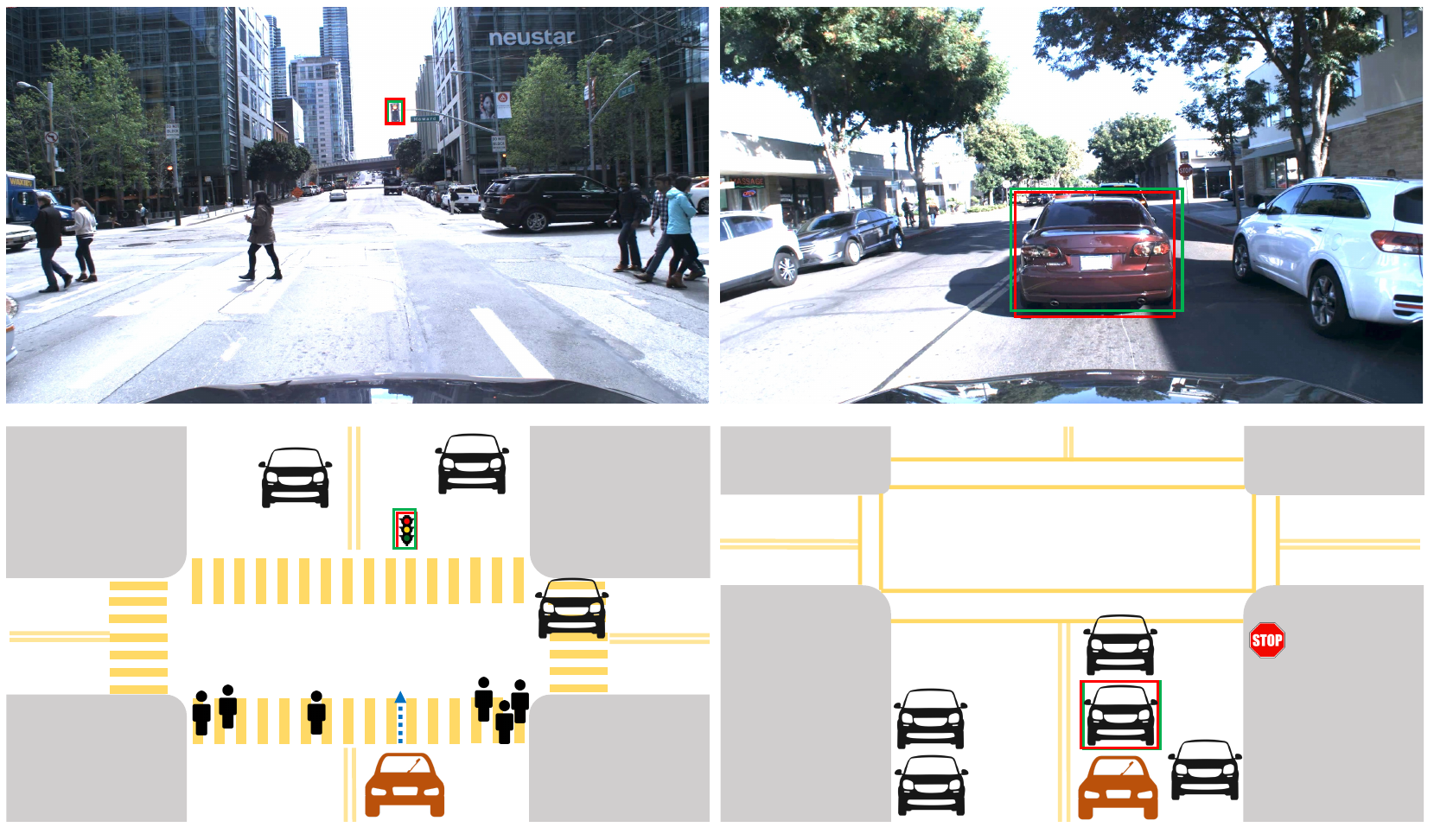}
      \end{subfigure}
    
      \vspace{0.5em}
    
      \begin{subfigure}[b]{0.45\textwidth}
        \centering
        \includegraphics[width=\linewidth,height=3.6cm]{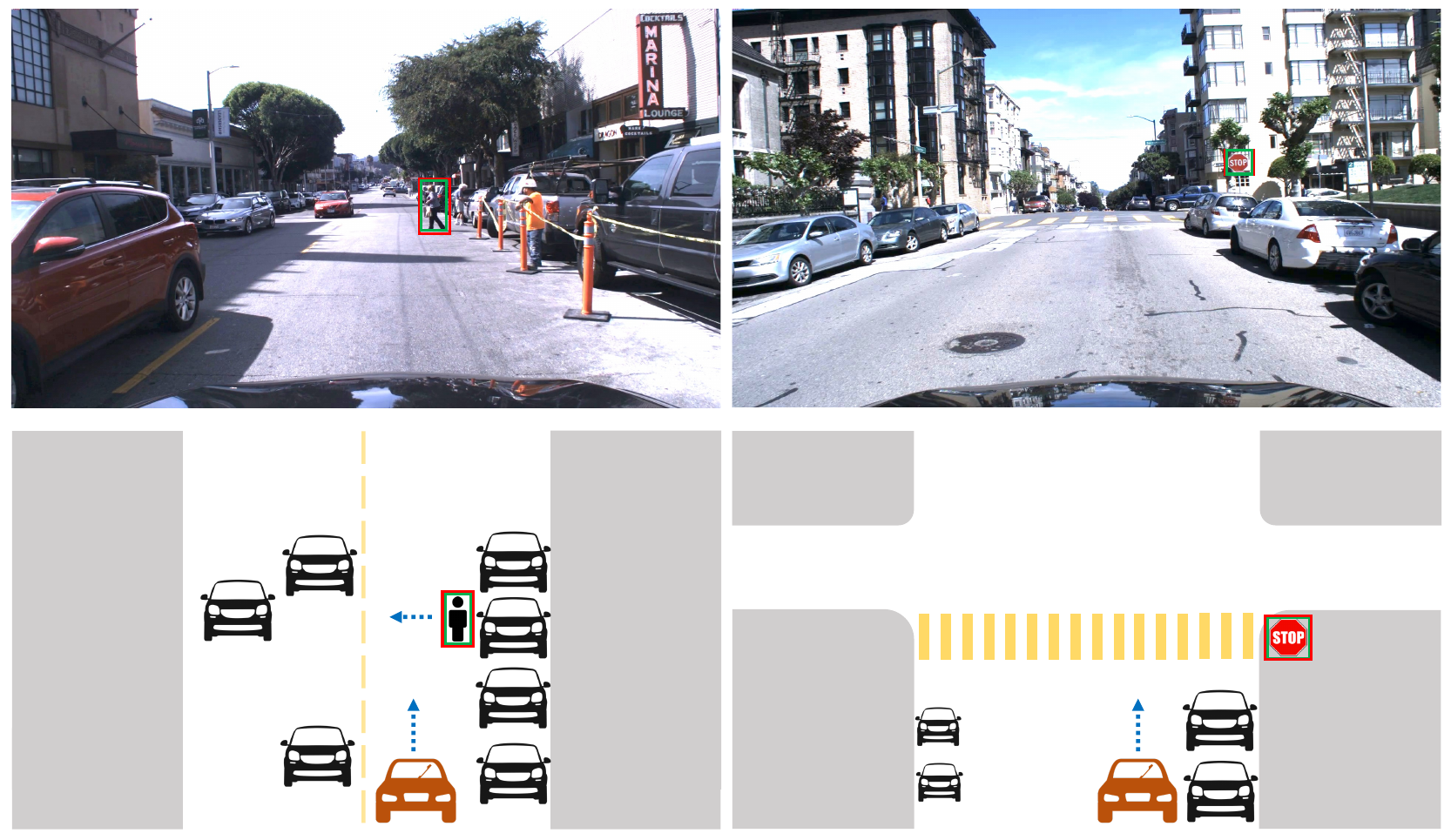}
      \end{subfigure}
      \begin{subfigure}[b]{0.45\textwidth}
        \centering
        \includegraphics[width=\linewidth,height=3.6cm]{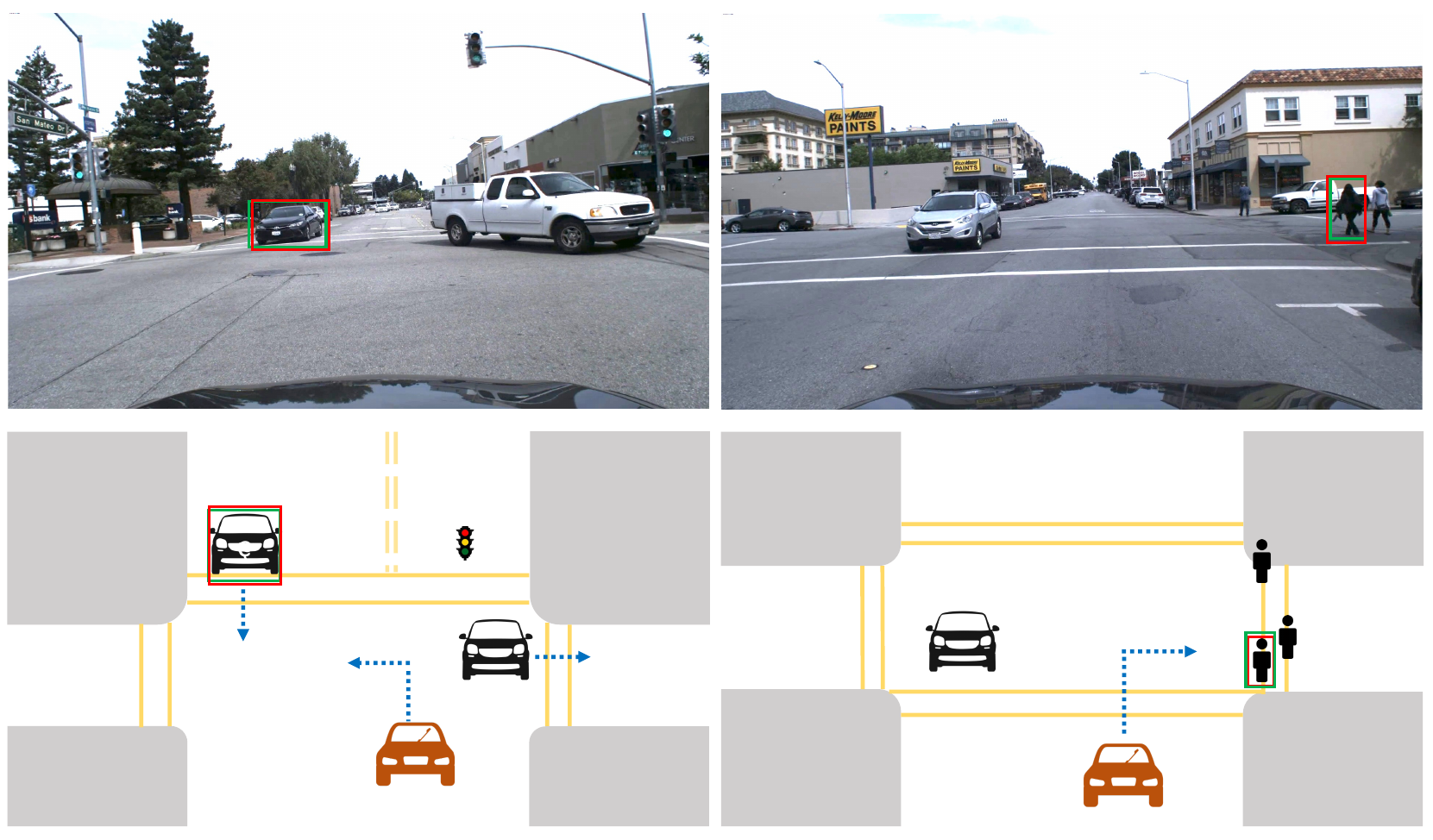}
      \end{subfigure}
    
      \caption{\textbf{Risk object identification results on the RAID dataset.} The ground truth is shown in red boxes, predictions are shown using green boxes. The ego-vehicle is depicted by the orange car, and blue arrow shows future motion direction. A birds-eye-view representation is presented below each front-view image providing information including scene layout, intentions of traffic participants, and the ego vehicle.}
      \label{fig:roionraid}
    \end{figure*}

    \begin{figure}[t!]
        \centering
        \includegraphics[width=0.49\textwidth,height=4.5cm]{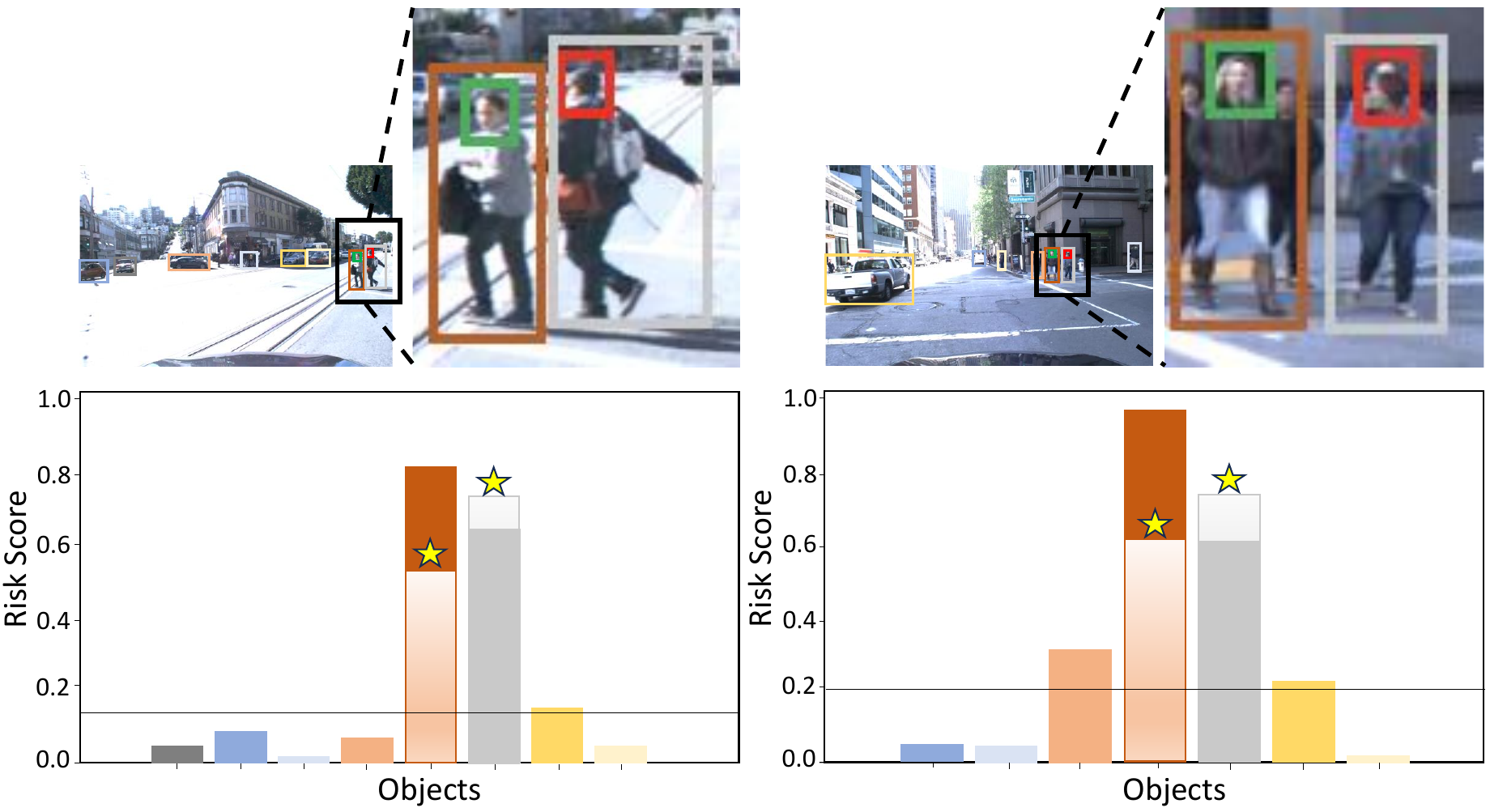}
        \caption{\textbf{Joint risk assessment on RAID}. Top row shows detected agents with colored boxes and bottom bar chart uses matching box colors for risk scores. The black line marks the predicted ‘Continue’ score without intervention, and $\star$ shows adjusted scores after factoring in pedestrian attention.}
        \label{fig:risk_qual}
        \vspace{-1em}
    \end{figure}
    
    \xhdr{Quantitative Results.} Table~\ref{comparison_hdd} shows that our base model (Ours) outperforms all baselines on the HDDS dataset~\cite{ Li_risk_object_IROS2020, li2023TPAMI}, and this improvement also generalizes well across different categories. Our base model even outperforms ~\cite{li2023TPAMI} which uses driver action for supervision, demonstrating the efficacy of our graph based modeling. Note that since the driver action labels from HDDS are not publicly released, we are unable to incorporate driver action in our model in Table~\ref{comparison_hdd}. 
    
    We see a similar trend of our base model outperforming prior state of the art on the proposed RAID dataset in Table~\ref{comparison_roi}. When we add the driver action module to our framework (Ours+), we see further improvement in results. However, in the case of \textit{Car Blocking Ego Lane}, performance notably drops—likely due to a label-behavior mismatch: Ours+ uses high-level driver action (e.g., \textit{Right-Turn}) as auxiliary supervision. In \textit{Car Blocking Ego Lane} scenarios, driver action is always labeled as \textit{Go-Straight} to reflect the intended route. However, the ego vehicle often performs lateral maneuvers to bypass the blockage—behavior that visually resembles a turn. Under weak supervision, this conflict may hinder the model’s ability to correctly associate such behavior with the blocking object, leading to misidentification. We plan to examine this effect more systematically in future work. 
    We also observe that traffic lights and stop signs perform noticeably worse compared to other categories. This is mainly due to i) inconsistent detection and tracking and ii) added challenge of reasoning about objects that are not directly in the ego-vehicle’s path, requiring deeper scene understanding. Improved object detectors, stronger backbones, and more robust tracking algorithms could further enhance performance in these categories.
    Some results are visualized in Figure~\ref{fig:roionraid}.
    
    Beyond identifying risk objects, our framework also predicts the driver's action and response, as shown in Table~\ref{tab:other}. Incorporating the driver's action improves both driver response prediction and risk object identification, reinforcing our hypothesis that modeling driver response is effective for weakly supervised risk object identification. Interestingly, left turns under driver's future action show significantly lower performance than right turns, due to their higher contextual complexity—such as unprotected intersections and oncoming traffic—which adds spatial and temporal variability. Right turns, being more structured and often free, are easier to model.
    %
    %
    
    Table~\ref{ped_attn} presents classification results for pedestrian attention, showing that using face inputs significantly outperforms~\cite{Rasouli_ICCVW2017}, as faces provide stronger attention cues. Both methods use the same ResNet-101 backbone for fair comparison, differing only in the training boxes (face vs. body). Performance is higher for the \textit{Not Looking} class, largely due to its greater representation in the training data—a reflection of natural real-world distribution since pedestrians rarely sustain eye contact, a temporal imbalance well-documented in transportation research (e.g., NHTSA, urban mobility studies).
    We also report average precision for pedestrian attention detection at an IoU threshold of 0.5. In practical settings, detection is more relevant than classification, as ground truth bounding boxes are typically unavailable. Interestingly, \textit{Not Looking} performs worse in detection despite having more training samples. This is likely because faces looking away are harder to detect, especially in driving scenes where faces are often small—unlike classification, which benefits from annotated boxes.
    

    \xhdr{Qualitative Results.}  The presence or absence of a road agent’s attention towards the ego-vehicle can influence—but not eliminate—the level of risk they pose. For example, a jaywalker who is attentive towards ego-vehicle still represents a risk, albeit a reduced one. Figure~\ref{fig:risk_qual} illustrates this using two scenarios, where the bar chart depicts the risk score of each traffic agent using \textit{only} the risk object identification framework. Two crossing pedestrians are highlighted in each scenario; one initially has a higher risk score, but their attentiveness towards the ego-vehicle lowers their final risk score (Eq.~\ref{joint_risk}), denoted by $\star$. We hope to further investigate the degree of pedestrian attention's effect on risk perception in the future, but in general, we observe that attention definitely contributes to driver's risk perception. 

    \section{Conclusion}
    We study risk perception in driving scenes by introducing a new dataset and a weakly supervised framework for identifying risk objects that influence driver behavior. The dataset captures highly interactive traffic scenarios, including pedestrian attention, with annotated head positions and an attention detection module to analyze its impact on driver risk assessment. We are the first to model risk perception as the interplay between driver responses and pedestrian attentiveness, advancing human–AI systems toward more holistic risk anticipation. Future work will incorporate road topology to further improve performance.

	
	\bibliographystyle{IEEEtran}
	\bibliography{root} 

@String(PAMI = {IEEE Trans. Pattern Anal. Mach. Intell.})

@String(CVPR= {IEEE Conf. Comput. Vis. Pattern Recog.})

@String(ICCV= {Int. Conf. Comput. Vis.})

@String(ECCV= {Eur. Conf. Comput. Vis.})

@String(BMVC= {Brit. Mach. Vis. Conf.})

@String(ICIP = {IEEE Int. Conf. Image Process.})

@String(ACCV  = {ACCV})

@String(ICLR = {Int. Conf. Learn. Represent.})

@String(PAMI  = {IEEE TPAMI})

@String(CVPR  = {CVPR})

@String(ICCV  = {ICCV})

@String(ECCV  = {ECCV})

@String(BMVC  =	{BMVC})

@String(ICIP  = {ICIP})

@String(ICLR  = {ICLR})

@misc{who,
author = {{World Health Organization}},
title = {{Global Status Report on Road Safety 2018: Summary}},
year = {2018},
}

@article{Wang_DriverBeahvior2014,
author = {Wenshuo Wang and Junqiang Xi and Huiyan Chen},
title = {{Modeling and Recognizing Driver Behavior Based on Driving Data: A Survey}},
journal = {Mathematical Problems in Engineering},
year = 2014
}

@article{li2023TPAMI,
author = {Chengxi Li and Stanley H Chan and Yi-Ting Chen},
title={{DROID: Driver-centric Risk Object Identification}},
journal = {PAMI},
year = {2023}
}

@article{Rasouli_review_ITS2020,
author = {Amir Rasouli and John K Tsotsos},
title = {{Autonomous Vehicles that Interact with Pedestrians: A Survey of Theory and Practice}},
journal = {ITS},
year = {2020}
}

@inproceedings{Li_risk_object_IROS2020,
    author={Chengxi Li and Stanley H. Chan and Yi-Ting Chen},
    title={{Who Make Drivers Stop? Towards Driver-centric Risk Assessment: Risk Object Identification via Causal Inference}},
    booktitle={IROS},
    year={2020}
}

@inproceedings{Rasouli_ICCVW2017,
    author={Amir Rasouli and Iuliia Kotseruba and John K. Tsotsos},
    title={{Are They Going to Cross? A Benchmark Dataset and Baseline for Pedestrian Crosswalk Behavior}},
    booktitle={ICCV-W},
    year={2017},
}

@inproceedings{Rasouli_pie_ICCV2019,
    author={Amir Rasouli and Iuliia Kotseruba and Toni Kunic and John K Tsotsos},
    title={{PIE: A Large-scale Dataset and Models for Pedestrian Intention Estimation and Trajectory Prediction}},
    booktitle={ICCV},
    year={2019},
}

@article{risk_assessment_Lefevre_ROBOMECH,
	Author = {Lef{\`e}vre, St{\'e}phanie and Vasquez, Dizan and Laugier, Christian},
    Title = {{A Survey on Motion Prediction and Risk Assessment for Intelligent Vehicles}},
	Journal = {ROBOMECH Journal},
	Year = {2014}
	}

@inproceedings{Kooij_pedestrianContext_ECCV2014,
    author={J. Kooij and N. Schneider and F. Flohr and D. Gavrila},
    title={{Context-based Pedestrian Path Prediction}},
    booktitle={ECCV},
    year={2014},
}

@inproceedings{tawari2018learning,
  title={{Learning to Attend to Salient Targets in Driving Videos using Fully Convolutional RNN}},
  author={Tawari, Ashish and Mallela, Praneeta and Martin, Sujitha},
  booktitle={ITSC},
  year={2018}
}

@inproceedings{wang2019deep,
  title={{Deep Object-centric Policies for Autonomous Driving}},
  author={Wang, Dequan and Devin, Coline and Cai, Qi-Zhi and Yu, Fisher and Darrell, Trevor},
  booktitle={ICRA},
  year={2019}
}

@inproceedings{kim2017interpretable,
  title={{Interpretable Learning for Self-driving Cars by Visualizing Causal Attention}},
  author={Kim, Jinkyu and Canny, John},
  booktitle={ICCV},
  year={2017}
}

@inproceedings{deng2019retinaface,
  title={{Retinaface: Single-stage Dense Face Localisation in the Wild}},
  author={Deng, Jiankang and Guo, Jia and Zhou, Yuxiang and Yu, Jinke and Kotsia, Irene and Zafeiriou, Stefanos},
  booktitle={CVPR},
  year={2020}
}

@inproceedings{Zeng_agentrisk_cvpr2017,
  title={{Agent-Centric Risk Assessment: Accident Anticipation and Risky Region Localization}},
  author={Kuo-Hao Zeng and Shih-Han Chou and Fu-Hsiang Chan and Juan Carlos Niebles and Min Sun},
  booktitle={CVPR},
  year={2017}
}

@inproceedings{Alletto_Dreye_cvprw2016,
  title={{DR(eye)VE: A Dataset for Attention-based Tasks with Applications to Autonomous and Assisted Driving}},
  author={Stefano Alletto and Andrea Palazzi and Francesco Solera and Simone Calderara and Rita Cucchiara},
  booktitle={CVPR-W},
  year={2016}
}

@inproceedings{Haan_causalconfusion_nips2019,
  title={{Causal Confusion in Imitation Learning}},
  author={Pim de Haan and Dinesh Jayaraman and Sergey Levine},
  booktitle={NeuIPS},
  year={2019}
}

@inproceedings{Geiger2012CVPR,
  author = {Andreas Geiger and Philip Lenz and Raquel Urtasun},
  title = {{Are we ready for Autonomous Driving? The KITTI Vision Benchmark Suite}},
  booktitle = {CVPR},
  year = {2012}
}

@inproceedings{Wang_repulsion_cvpr2018,
  author = {Xinlong Wang and Tete Xiao and Yuning Jiang and Shuai Shao and Jian Sun and Chunhua Shen},
  title = {{Repulsion Loss: Detecting Pedestrians in a Crowd}},
  booktitle = {CVPR},
  year = {2018}
}

@inproceedings{Zhang_OR_eccv_2018,
  author = {Shifeng Zhang and Longyin Wen and Xiao Bian and Zhen Lei and Stan Z. Li},
  title = {{Occlusion-aware R-CNN: Detecting Pedestrians in a Crowd}},
  booktitle = {ECCV},
  year = {2018}
}

@inproceedings{Zhang_filteredchannel_cvpr2015,
  author = {Shanshan Zhang and Rodrigo Benenson and Bernt Schiele},
  title = {{Filtered Channel Features for Pedestrian Detection}},
  booktitle = {CVPR},
  year = {2015}
}

@inproceedings{luo_multimodal_cvpr2020,
  author = {Yan Luo and Chongyang Zhang and Muming Zhao and Hao Zhou and Jun Sun},
  title = {{Where, What, Whether: Multi-modal Learning Meets Pedestrian Detection}},
  booktitle = {CVPR},
  year = {2020}
}

@inproceedings{Malla_TITAn_cvpr2020,
  author = {Srikanth Malla and Behzad Dariush and Chiho Choi},
  title = {{TITAN: Future Forecast using Action Priors}},
  booktitle = {CVPR},
  year = {2020}
}

@inproceedings{Alahi_sociallstm_cvpr2016,
  author = {Alexandre Alahi and Kratarth Goel and Vignesh Ramanathan and Alexandre Robicquet and Li Fei-Fei and Silvio Savarese},
  title = {{Social LSTM: Human Trajectory Prediction in Crowded Spaces}},
  booktitle = {CVPR},
  year = {2016}
}

@inproceedings{Gupta_socialgan_cvpr2018,
  author = {Agrim Gupta and Justin Johnson and Li Fei-Fei and Silvio Savarese and Alexandre Alahi},
  title = {{Social GAN: Socially Acceptable Trajectories with Generative Adversarial Networks}},
  booktitle = {CVPR},
  year = {2018}
}

@inproceedings{Bera_adapt_icra2014,
  author = {Aniket Bera and Nico Galoppo and Dillon Sharlet and Adam Lake and Dinesh Manocha},
  title = {{AdaPT: Real-time Adaptive Pedestrian Tracking for Crowded Scenes}},
  booktitle = {ICRA},
  year = {2014}
}

@inproceedings{Wojke2018deep,
  title={{Deep Cosine Metric Learning for Person Re-identification}},
  author={Wojke, Nicolai and Bewley, Alex},
  booktitle={WACV},
  year={2018}
}

@inproceedings{Bergmann_tracking_iccv2019,
  title={{Tracking without bells and whistles}},
  author={Philipp Bergmann and Tim Meinhardt and Laura Leal-Taixe},
  booktitle={ICCV},
  year={2019}
}

@inproceedings{mot16,
  title={{MOT16: A Benchmark for Multi-Object Tracking}},
  author={Anton Milan and Laura Leal-Taixe and Ian Reid and Stefan Roth and Konrad Schindler},
  booktitle={arXiv:1603.00831},
  year={2016}
}

@inproceedings{Rezatofighi_iccv2015,
  title={{Joint Probabilistic Data Association Revisited}},
  author={Seyed Hamid Rezatofighi and Anton Milan and Zhen Zhang and Qinfeng Shi and Anthony Dick and Ian Reid},
  booktitle={ICCV},
  year={2015}
}

@inproceedings{Rasouli_pedetrianaction_bmvc2018,
  title={{Pedestrian Action Anticipation using
Contextual Feature Fusion in Stacked RNNs}},
  author={Amir Rasouli and Iuliia Kotseruba and John K. Tsotsos},
  booktitle={BMVC},
  year={2019}
}

@inproceedings{Hamaoka_headturning_ivw2013,
  title={{A Study on the Behavior of Pedestrians when Confirming Approach of Right/Left-turning Vehicle while Crossing a Crosswalk}},
  author={Hidekatsu Hamaoka and Toru Hagiwara and Masahiro Tada and Kazunori Munehiro},
  booktitle={IV-W},
  year={2013}
}

@inproceedings{he2016deep,
  title={{Deep Residual Learning for Image Recognition}},
  author={He, Kaiming and Zhang, Xiangyu and Ren, Shaoqing and Sun, Jian},
  booktitle={CVPR},
  year={2016}
}

@inproceedings{he2017mask,
  title={{Mask R-CNN}},
  author={He, Kaiming and Gkioxari, Georgia and Doll{\'a}r, Piotr and Girshick, Ross},
  booktitle={ICCV},
  year={2017}
}

@inproceedings{wojke2017simple,
  title={{Simple Online and Realtime Tracking with a Deep Association Metric}},
  author={Wojke, Nicolai and Bewley, Alex and Paulus, Dietrich},
  booktitle={ICIP},
  year={2017}
}

@article{kipf2016semi,
  title={{Semi-supervised Classification with Graph Convolutional Networks}},
  author={Kipf, Thomas N and Welling, Max},
  journal={ICLR},
  year={2017}
}

@article{hochreiter1997long,
  title={Long short-term memory},
  author={Hochreiter, Sepp and Schmidhuber, J{\"u}rgen},
  journal={Neural computation},
  year={1997}
}

@inproceedings{liu2018image,
  title={{Image Inpainting for Irregular Holes using Partial Convolutions}},
  author={Liu, Guilin and Reda, Fitsum A and Shih, Kevin J and Wang, Ting-Chun and Tao, Andrew and Catanzaro, Bryan},
  booktitle={ECCV},
  year={2018}
}

@inproceedings{Endsley_SA_2000,
  title={{Theoretical Underpinnings of Situation Awareness: A
Critical Review}},
  author={Mica R. Endsley},
  booktitle={Situation awareness analysis},
  year={2000}
}

@inproceedings{zhang2019self,
  title={A Self Validation Network for Object-Level Human Attention Estimation},
  author={Zhang, Zehua and Yu, Chen and Crandall, David},
  booktitle={NeurIPS},
  year={2019}
}

@inproceedings{porzi2019seamless,
  title={{Seamless Scene Segmentation}},
  author={Porzi, Lorenzo and Bulo, Samuel Rota and Colovic, Aleksander and Kontschieder, Peter},
  booktitle={CVPR},
  year={2019}
}

@inproceedings{yang2016wider,
  title={{Wider face: A Face Detection Benchmark}},
  author={Yang, Shuo and Luo, Ping and Loy, Chen-Change and Tang, Xiaoou},
  booktitle={CVPR},
  year={2016}
}

@inproceedings{lin2014microsoft,
  title={{Microsoft COCO: Common Objects in Context}},
  author={Lin, Tsung-Yi and Maire, Michael and Belongie, Serge and Hays, James and Perona, Pietro and Ramanan, Deva and Doll{\'a}r, Piotr and Zitnick, C Lawrence},
  booktitle={ECCV},
  year={2014}
}

@article{riskperception,
  title={Quantitative Assessment of Driver's risk Perception Using a Simulator},
  author={Mitsuteru Kokubun and Hiroyuki Konishi and Kazunori Higuchi and Tetsuo Kurahashi and Yoshiyuki Umemura and Hiroaki Nishi},
  journal={International Journal of Vehicle Safety},
  year={2005}
}

@inproceedings{girase2021loki,
  title={LOKI: Long Term and Key Intentions for Trajectory Prediction},
  author={Girase, Harshayu and Gang, Haiming and Malla, Srikanth and Li, Jiachen and Kanehara, Akira and Mangalam, Karttikeya and Choi, Chiho},
  booktitle={ICCV},
  year={2021}
}

@article{liu2020spatiotemporal,
  title={Spatiotemporal relationship reasoning for pedestrian intent prediction},
  author={Liu, Bingbin and Adeli, Ehsan and Cao, Zhangjie and Lee, Kuan-Hui and Shenoi, Abhijeet and Gaidon, Adrien and Niebles, Juan Carlos},
  journal={RA-L},
  year={2020}
}

@inproceedings{li2022important,
  title={Important Object Identification with Semi-Supervised Learning for Autonomous Driving},
  author={Li, Jiachen and Gang, Haiming and Ma, Hengbo and Tomizuka, Masayoshi and Choi, Chiho},
  booktitle={ICRA},
  year={2022}
}

@inproceedings{gao2019goal,
  title={Goal-oriented object importance estimation in on-road driving videos},
  author={Gao, Mingfei and Tawari, Ashish and Martin, Sujitha},
  booktitle={ICRA},
  year={2019}
}

@inproceedings{wu2019learning,
  title={Learning actor relation graphs for group activity recognition},
  author={Wu, Jianchao and Wang, Limin and Wang, Li and Guo, Jie and Wu, Gangshan},
  booktitle={CVPR},
  year={2019}
}

@article{shrout1979intraclass,
  title={Intraclass correlations: uses in assessing rater reliability.},
  author={Shrout, Patrick E and Fleiss, Joseph L},
  journal={Psychological bulletin},
  year={1979}
}

@article{cicchetti1994guidelines,
  title={Guidelines, criteria, and rules of thumb for evaluating normed and standardized assessment instruments in psychology.},
  author={Cicchetti, Domenic V},
  journal={Psychological assessment},
  year={1994}
}

@inproceedings{sachdeva2024rank2tell,
  title={Rank2tell: A multimodal driving dataset for joint importance ranking and reasoning},
  author={Sachdeva, Enna and Agarwal, Nakul and Chundi, Suhas and Roelofs, Sean and Li, Jiachen and Kochenderfer, Mykel and Choi, Chiho and Dariush, Behzad},
  booktitle={WACV},
  year={2024}
}

@inproceedings{malla2023drama,
  title={Drama: Joint risk localization and captioning in driving},
  author={Malla, Srikanth and Choi, Chiho and Dwivedi, Isht and Choi, Joon Hee and Li, Jiachen},
  booktitle={WACV},
  year={2023}
}

@article{chung2014empirical,
  title={Empirical evaluation of gated recurrent neural networks on sequence modeling},
  author={Chung, Junyoung and Gulcehre, Caglar and Cho, KyungHyun and Bengio, Yoshua},
  journal={arXiv preprint arXiv:1412.3555},
  year={2014}
}

@inproceedings{lea2017temporal,
  title={Temporal convolutional networks for action segmentation and detection},
  author={Lea, Colin and Flynn, Michael D and Vidal, Rene and Reiter, Austin and Hager, Gregory D},
  booktitle={CVPR},
  year={2017}
}

@article{dax2023disentangled,
  title={Disentangled neural relational inference for interpretable motion prediction},
  author={Dax, Victoria M and Li, Jiachen and Sachdeva, Enna and Agarwal, Nakul and Kochenderfer, Mykel J},
  journal={RA-L},
  year={2023}
}

@inproceedings{xia2019predicting,
  title={Predicting driver attention in critical situations},
  author={Xia, Ye and Zhang, Danqing and Kim, Jinkyu and Nakayama, Ken and Zipser, Karl and Whitney, David},
  booktitle={ACCV},
  year={2018}
}

@inproceedings{agarwal2023ordered,
  title={Ordered atomic activity for fine-grained interactive traffic scenario understanding},
  author={Agarwal, Nakul and Chen, Yi-Ting},
  booktitle={ICCV},
  year={2023}
}

@inproceedings{chi2023adamsformer,
  title={Adamsformer for spatial action localization in the future},
  author={Chi, Hyung-gun and Lee, Kwonjoon and Agarwal, Nakul and Xu, Yi and Ramani, Karthik and Choi, Chiho},
  booktitle={CVPR},
  year={2023}
}

@misc{agarwal2023driver,
  title={Driver behavior risk assessment and pedestrian awareness},
  author={Agarwal, Nakul and Chen, Yi-Ting},
  year={2023},
  publisher={Google Patents},
  note={US Patent 11,845,464}
}
	
\end{document}